\documentclass[runningheads]{llncs}
\usepackage[T1]{fontenc}
\usepackage{latexsym}
\usepackage{graphicx}
\usepackage{amsmath}
\usepackage[hidelinks]{hyperref}
\usepackage{booktabs}
\usepackage{caption}
\usepackage{orcidlink}
\usepackage{algorithm}
\usepackage{hyperref} 
\hypersetup{hidelinks}
\usepackage{algorithmic}
\usepackage[dvipsnames]{xcolor}
\usepackage{subcaption}
\usepackage{arydshln}
\usepackage{todonotes}
\usepackage{float}
\usepackage{microtype}
\usepackage{booktabs}
\usepackage{caption} 
\usepackage{colortbl}
\usepackage{xcolor}
\usepackage[most]{tcolorbox}
\tcbuselibrary{listings,breakable,skins}
\usepackage{xcolor}
\usepackage{marvosym}

\definecolor{promptpurple}{HTML}{7D78B8}
\definecolor{promptbg}{HTML}{FAFAFF}

\newtcblisting[auto counter, number within=section]{promptlisting}[2][]{
  enhanced,
  breakable,
  listing only,
  colback=promptbg,
  colframe=promptpurple,
  coltitle=white,
  colbacktitle=promptpurple,
  title={Prompt~\thetcbcounter: #2},
  label={#1},
  fonttitle=\bfseries\small,
  boxrule=1pt,
  arc=0pt,
  outer arc=0pt,
  left=6pt,
  right=6pt,
  top=6pt,
  bottom=6pt,
  listing options={
    basicstyle=\ttfamily\scriptsize,
    breaklines=true,
    columns=fullflexible,
    keepspaces=true,
    showstringspaces=false
  }
}
\begin{document}
\title{\textsc{Graph2Idea}: Retrieval-Augmented Scientific Idea Generation with Graph-Structured Contexts}
\titlerunning{Graph2Idea}
%
\author{Xu Li\inst{1}\textsuperscript{\href{mailto:xul@swpu.edu.cn}{(\Letter)}}\orcidlink{0000-0002-5725-2677}
\and
Hanzhe Tu\inst{2}\orcidlink{0009-0006-8476-4826} 
Xun Han\inst{3}\textsuperscript{\href{mailto:hldwxhx@163.com}{(\Letter)}}
}

\authorrunning{Xu Li, et al.}
%
\institute{Southwest Petroleum University, Chengdu, Sichuan, China\\
\email{\{xul\}@swpu.edu.cn},
\email{\{tuhanzhe.swust\}@qq.com}
\and
Sichuan Police College, Luzhou, Sichuan, China\\
\email{\{hldwxhx\}@163.com}
}

%
\maketitle              

\begin{abstract}
Generating novel, feasible, and high-quality research ideas is an important yet challenging task in scientific discovery.
Recent Large Language Model (LLM)-based methods often ground idea generation with retrieved literature, but the retrieved evidence is usually provided as flat text, such as titles, abstracts, or summaries. Such flat contexts may contain redundant or weakly relevant information, while making cross-paper relations among problems, methods, mechanisms, and findings difficult to identify and trace.
To address this challenge, we propose \textsc{Graph2Idea}, a knowledge graph-guided framework for retrieval-augmented scientific idea generation.
\textsc{Graph2Idea} first retrieves papers according to the input topic, transforms them into structured knowledge triples, and dynamically constructs a target-centered knowledge graph to make literature relations explicit.
It then extracts compact graph-derived contexts that retain target-relevant relational evidence while reducing noisy textual input.
Based on these contexts, a two-stage generation process first identifies promising research directions and then guides the LLM to synthesize candidate ideas from graph-grounded evidence.
Experiments on a scientific idea generation benchmark show that \textsc{Graph2Idea} outperforms representative baselines under the automatic evaluation protocol.
Compared with the strongest baseline scores, it improves \textit{Novelty} from 0.45 to 0.52, \textit{Quality} from 0.24 to 0.29, and \textit{Feasibility} from 0.22 to 0.28.
These results suggest that graph-structured evidence helps LLMs generate research ideas through more explicit, compact, and traceable recombination of prior scientific knowledge.

\keywords{Scientific idea generation \and Knowledge graph \and Retrieval-augmented generation \and Large language models \and Scientific knowledge recombination.}
\end{abstract}
\section{Introduction}
Generating novel and meaningful research ideas is a central cognitive activity in scientific research \cite{baek_researchagent_2025}. As a fundamental task in AI for Science, scientific idea generation aims to assist researchers in identifying promising research directions from rapidly growing scientific literature \cite{shahhosseini2026largelanguagemodelsscientific}. Researchers typically synthesize large volumes of previous work, understand existing research directions, identify unresolved problems, and envision new methodological or conceptual advances \cite{hope2023computationalinflectionscientificdiscovery,Wang2023ScientificDI}. This process is inherently time-consuming and intellectually demanding, and has become increasingly challenging with the rapid growth and fragmentation of scientific literature \cite{fire2018overoptimizationacademicpublishingmetrics}. Recent advances in Large Language Models (LLMs) have demonstrated strong capabilities for knowledge synthesis by integrating effective reasoning and information retrieval techniques, suggesting promising opportunities to support literature-intensive scientific idea generation \cite{openai2024gpt4technicalreport,ma_llm_2024,romera-paredes_mathematical_2024,trivedi2023interleavingretrievalchainofthoughtreasoning}.

LLM-assisted scientific idea generation frameworks are commonly built upon the Retrieval-Augmented Generation (RAG) paradigm \cite{lewis2021retrievalaugmentedgenerationknowledgeintensivenlp,10.1007/978-981-95-4088-4_16}, which uses external academic knowledge to contextualize generated concepts. Early studies mainly focused on retrieval optimization, such as iterative semantic search or relevance-based literature retrieval, to improve the grounding and novelty of generated ideas \cite{baek_researchagent_2025,si_can_2024}. More recent methods have introduced increasingly complex generation strategies, including planned retrieval for expanded topical coverage \cite{hu_nova_2024}, chain-based modeling of research trends \cite{li_chain_2024}, prompting-based reasoning \cite{wei2023chainofthoughtpromptingelicitsreasoning,meincke_prompting_2024}, and multi-agent collaboration \cite{su_many_2025,chen_enhancing_nodate}.

Despite these advances, most existing retrieval-augmented approaches still provide retrieved literature as flat textual evidence, such as paper titles, abstracts, or summaries. Although such contexts contain useful background knowledge, they may also introduce retrieval noise, repeated descriptions, and weakly relevant information. More importantly, flat textual contexts do not explicitly represent how scientific problems, methods, mechanisms, findings, and limitations are related across different studies. Without such explicit relational organization, LLMs have to identify useful evidence, filter irrelevant content, and infer possible recombination paths directly from unstructured text, which may weaken the controllability and traceability of scientific idea generation. 
This limitation is particularly important because many valuable research ideas do not arise from unconstrained semantic divergence alone, but from meaningful recombination of existing knowledge components \cite{KGR,sternlicht2026chimeraknowledgebasescientific}. For example, a new idea may transfer a method to a different problem setting, combine complementary mechanisms from different studies, or reorganize known findings into a new experimental direction. Such recombinational idea generation requires not only relevant literature, but also a compact and explicit representation of how prior knowledge can be connected and reused.

To address this limitation, we propose \textsc{Graph2Idea}, a knowledge graph-guided framework for retrieval-augmented scientific idea generation. Given a target paper, \textsc{Graph2Idea} first extracts its research intent, including its purpose, mechanism, method summary, and keywords, and uses this information to retrieve relevant studies. Instead of directly feeding retrieved titles and abstracts to an LLM, \textsc{Graph2Idea} extracts structured knowledge triples from paper titles and abstracts, and dynamically constructs a target-centered knowledge graph that captures relational connections among scientific knowledge components. The graph is then used to build compact graph-derived contexts, which provide denser, more structured, and more traceable evidence for idea generation. Based on these graph-derived contexts, \textsc{Graph2Idea} further decomposes idea generation into two stages: the LLM first identifies promising research directions under different generation strategies, and then synthesizes candidate research ideas conditioned on the selected directions and graph-grounded evidence. This two-stage design separates direction planning from idea synthesis, allowing the model to recombine structured literature knowledge in a more controllable manner. Taken together, \textsc{Graph2Idea} aims to support early-stage scientific idea generation by generating novel, feasible, and high-quality research ideas through explicit recombination of graph-structured literature evidence, rather than relying solely on noisy flat retrieved text.

Our contributions are summarized as follows:

\begin{itemize}
    \item We propose \textsc{Graph2Idea}, a knowledge graph-guided framework that converts retrieved literature into structured graph-derived contexts for scientific idea generation.


    \item We introduce a two-stage generation process that first plans research directions and then synthesizes candidate ideas, supporting controllable recombination of graph-grounded knowledge.

    \item We evaluate \textsc{Graph2Idea} on a scientific idea generation benchmark. Results show improvements in novelty, feasibility, and quality over representative baselines.
\end{itemize}

\section{Related Work}
\subsection{LLM-Driven Idea Generation}

Recent advances in LLMs have sparked growing interest in AI-assisted scientific idea generation within the broader AI for Science paradigm
\cite{tang2025airesearcherautonomousscientificinnovation,luo2025llm4srsurveylargelanguage}.
Most existing work adopts retrieval-augmented generation as the core approach, using external academic knowledge to ground the generation of new research ideas.
Representative methods construct retrieval mechanisms based on semantic similarity, entity associations, or citation-based signals to identify relevant prior work for idea synthesis
\cite{baek_researchagent_2025,si_can_2024}.
Building upon this paradigm, subsequent studies introduce more structured retrieval and generation strategies, including planning-based retrieval to guide query formulation and expand topical coverage
\cite{hu_nova_2024,chen_enhancing_nodate}, chain-based modeling of research evolution \cite{li_chain_2024}, prompting-based reasoning
\cite{wei2023chainofthoughtpromptingelicitsreasoning,meincke_prompting_2024}, and multi-agent discussion settings
\cite{su_many_2025,lu_ai_2024}.

These studies show that LLMs can generate scientifically meaningful and sometimes highly novel research ideas when supported by retrieved literature and structured generation procedures
\cite{Girotra2023IdeasAD,kumar2025largelanguagemodelsunlock}.
However, most retrieval-augmented idea generation methods still provide retrieved papers as flat textual contexts, such as titles, abstracts, or summaries.
This makes the model responsible for filtering useful evidence and inferring cross-paper relations from unstructured text, which may limit the controllability and traceability of generated ideas.
For literature-grounded scientific idea generation, a useful idea often depends not only on individual papers, but also on how problems, methods, mechanisms, findings, and limitations can be connected across papers.
This motivates the use of structured knowledge representations that organize retrieved literature into compact graph-derived relational contexts.

\subsection{Structured Scientific Knowledge for Idea Generation}

Knowledge graphs provide a structured representation for organizing entities, concepts, and their relations, and have been widely used in knowledge-intensive tasks such as retrieval, question answering, and reasoning
\cite{Knowledge_survey,10.1007/978-981-95-4088-4_16}.
In scientific domains, prior studies have explored the construction of knowledge graphs from scholarly articles, citation networks, entity mentions, and extracted relational triples to support literature exploration and knowledge discovery
\cite{gao2025goaienhancingaistudents,baek_researchagent_2025}.
Recent graph-enhanced retrieval and knowledge-augmented LLM methods further suggest that relational structures can help LLMs access and use external knowledge more effectively in complex reasoning and generation tasks
\cite{peng2024graphretrievalaugmentedgenerationsurvey}.

For scientific idea generation, structured scientific knowledge can serve as a form of knowledge compression and relation organization.
Instead of exposing the LLM to long and potentially redundant paper abstracts, relevant information can be distilled into compact knowledge components and explicit relations.
Although knowledge graphs have been used for literature exploration and knowledge-intensive reasoning, their role in LLM-based scientific idea generation remains underexplored.
This suggests an opportunity to transform retrieved literature into graph-derived relational evidence, enabling more compact, traceable, and recombination-oriented idea generation.

\subsection{Planning-Guided LLM Generation}

The planning and reasoning capabilities of LLMs have become an active research focus for complex generation tasks.
Classical approaches such as Chain-of-Thought prompting~\cite{wei2023chainofthoughtpromptingelicitsreasoning} explicitly model intermediate reasoning steps, while Tree-of-Thought~\cite{yao2023treethoughtsdeliberateproblem} enables multi-branch exploration and self-evaluation.
For knowledge-intensive settings, frameworks such as ReAct~\cite{aksitov2023restmeetsreactselfimprovement} and Retrieval-Augmented Reasoning~\cite{cheng2025dualragdualprocessapproachintegrate} integrate reasoning, planning, and external knowledge access, showing that structured reasoning can improve complex generation.

Planning-based paradigms have also been explored in idea generation and hypothesis discovery tasks~\cite{hu_nova_2024,chen_enhancing_nodate}, where planning is often used to guide query formulation, organize intermediate reasoning, or support iterative refinement.
However, most existing methods apply planning directly over retrieved textual contexts, leaving the underlying knowledge structure implicit.
Planning over structured relational evidence can provide clearer intermediate guidance for idea generation, because the model can reason over explicit relations among problems, methods, mechanisms, and findings rather than only over unstructured retrieved text.
This motivates combining planning-guided generation with graph-derived contexts for more controllable scientific idea generation.

\section{Methodology}

\subsection{Graph2Idea Framework}

\begin{figure*}[t]
  \centering
  \includegraphics[width=\textwidth]{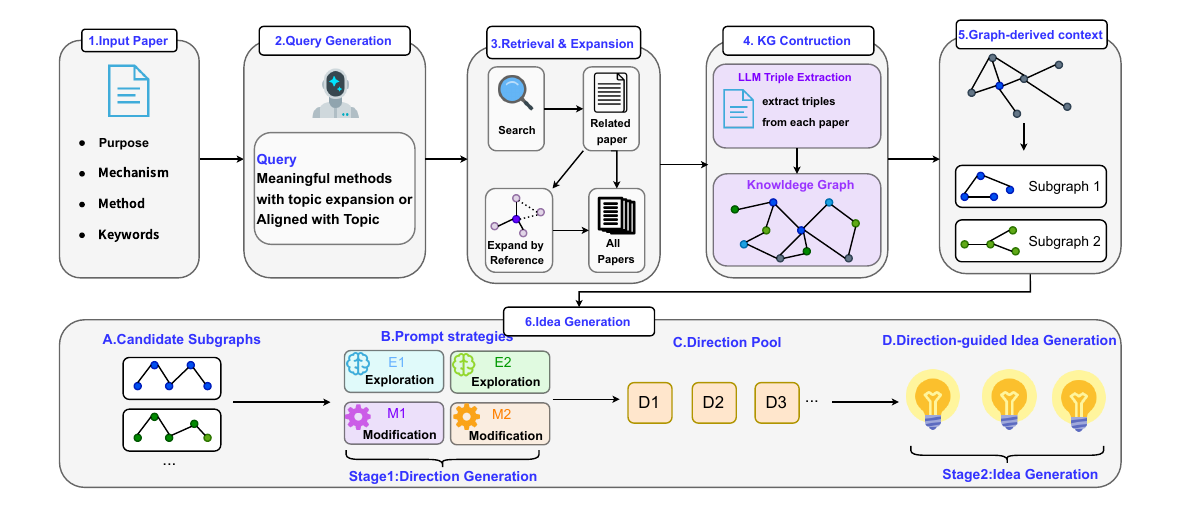}
  \caption{Knowledge graph-guided scientific idea generation framework.}

  \label{fig:framework}
\end{figure*}

Figure~\ref{fig:framework} illustrates the overall workflow of \textsc{Graph2Idea}.
Given an input paper, the framework proceeds through six pipeline steps grouped into four functional modules:
\textit{Target understanding and literature collection} (Steps 1--3),
\textit{Knowledge graph construction} (Step 4),
\textit{Graph-derived context construction} (Step 5),
and \textit{Two-stage idea generation} (Step 6).

Steps 1--3 extract a target profile, generate retrieval queries, and collect relevant literature.
Step 4 transforms the retrieved papers into structured knowledge triples and dynamically constructs a target-centered working graph.
Step 5 extracts compact graph-derived contexts that retain target-relevant relational evidence while reducing noisy flat textual input.
Step 6 generates candidate research ideas through direction planning and idea synthesis.
Overall, \textsc{Graph2Idea} converts retrieved literature from flat textual evidence into compact relational evidence for scientific idea generation.

\subsection{Target Understanding and Knowledge Graph Construction}

\textbf{Target understanding.} 
Given an input paper, an LLM first extracts a compact target profile, including its \textit{purpose}, \textit{mechanism}, \textit{method summary}, and \textit{keywords}. This profile defines the scope of scientific idea generation and guides the subsequent retrieval process. The LLM then generates multiple search queries that cover the target research problem, technical mechanism, related methods, and potentially transferable research directions.

\textbf{Literature collection.} 
To obtain a focused literature set for graph construction, \textsc{Graph2Idea} filters and expands the initially retrieved papers. Since the initial retrieval results may contain noisy or weakly relevant papers, we use an LLM-based ranker~\cite{si_can_2024} to select a small set of \textbf{base papers} according to their alignment with the target profile. The selected papers are further expanded through their references. Reference papers are ranked using signals such as keyword relevance, citation intent, publication year, abstract availability, and deduplication. The resulting literature set is used as the source for structured knowledge extraction rather than being directly concatenated as the final LLM input.

\textbf{Knowledge graph construction.} For each retained paper, an LLM extracts relational triples $(h,r,t)$ from its title and abstract, where $h$ and $t$ denote scientific knowledge components and $r$ denotes their relation. The extracted components may correspond to research problems, methods, mechanisms, findings, datasets, limitations, or application scenarios. To improve consistency, relation types are constrained by a predefined schema, such as \textit{addresses}, \textit{uses}, \textit{improves}, \textit{evaluated\_on}, and \textit{limited\_by}.

All triples are used to construct a directed knowledge graph $G=(V,E)$. For each triple, the head and tail are added as nodes, and the relation is added as a directed edge. We normalize node strings to reduce surface-form variation and merge duplicated triples with the same normalized head, relation, and tail while preserving source-paper metadata. In addition to semantic knowledge triples, the working graph also retains lightweight \textbf{auxiliary links}, such as publication year, venue, retrieval query, or source-paper information when available. These links are not treated as core scientific evidence; instead, they serve as bridge edges that connect otherwise sparse knowledge components during graph organization and context construction. As a result, the retrieved papers are transformed into compact relational knowledge units while preserving traceability to the original literature.

\subsection{Graph-Derived Context and Idea Generation}

\textbf{Graph-derived context construction.} 
Providing the entire working graph to the LLM would introduce noise and exceed the context budget. Therefore, \textsc{Graph2Idea} constructs multiple compact \textbf{graph-derived contexts}, as summarized in Algorithm~\ref{alg:subgraph}. Each context is built through two operations. First, \textbf{target-relevant seed selection} scores semantic edges according to their relevance to the target profile and selects high-scoring semantic edges as seed evidence. Second, \textbf{bridge-aware evidence expansion} expands each seed with related semantic triples that share core entities or can be reached through auxiliary bridge links. This allows auxiliary links to support evidence organization while preventing metadata-only relations from being directly used as LLM evidence.

During serialization, auxiliary metadata links are omitted from the final LLM context because they provide limited semantic evidence for idea generation. The serialized context mainly retains semantic triples among scientific problems, methods, mechanisms, findings, and limitations.Compared with flat retrieved abstracts, these graph-derived contexts provide more compact and focused relational evidence, allowing the LLM to observe how different papers may be connected through shared problems, methods, mechanisms, or source-related structures.

\begin{algorithm}[t]
\caption{Graph-Derived Context Construction}
\label{alg:subgraph}
\begin{algorithmic}[1]
\STATE \textbf{Input:} Working graph $G=(V,E)$, target profile $z$, number of contexts $M$, maximum edges $B$, maximum shared core nodes $\lambda$
\STATE \textbf{Output:} Serialized graph-derived contexts $\mathcal{C}$
\STATE Score semantic edges in $G$ according to their relevance to $z$.
\STATE Sort semantic edges in descending order of relevance score.
\STATE Initialize $\mathcal{E}_{used} \leftarrow \emptyset$, $\mathcal{V}_{core} \leftarrow \emptyset$, and $\mathcal{C} \leftarrow \emptyset$.
\FOR{$i = 1$ to $M$}
    \STATE Select high-scoring semantic seed edges not in $\mathcal{E}_{used}$ and satisfying the core-node overlap constraint $\lambda$.
    \IF{no seed edge is selected}
        \STATE break
    \ENDIF
    \STATE Expand each seed edge with related semantic triples using shared entities and auxiliary bridge links.
    \STATE Construct $G_i$ from the selected semantic triples.
    \STATE Update $\mathcal{E}_{used}$ and $\mathcal{V}_{core}$.
    \STATE Omit auxiliary metadata links and serialize semantic triples in $G_i$ into context $c_i$.
    \STATE $\mathcal{C} \leftarrow \mathcal{C} \cup \{c_i\}$.
\ENDFOR
\STATE \textbf{return} $\mathcal{C}$
\end{algorithmic}
\end{algorithm}

\textbf{Two-stage idea generation.} 
Given the graph-derived contexts, \textsc{Graph2Idea} generates candidate research ideas through a two-stage process. In the first stage, each context is paired with multiple prompt strategies, including exploratory recombination, mechanism transfer, guided improvement, and component reconfiguration. Under each strategy, the LLM identifies promising \textbf{research directions} that specify how graph evidence may be reused and what elements should be preserved or changed. In the second stage, each selected direction is used to guide final idea synthesis. Given the input paper, the graph-derived context, and the selected direction, the LLM generates a structured research idea containing a title, research problem, motivation, method, experiment design, expected contribution, and graph evidence. This design separates high-level direction planning from detailed idea formulation, making the generated ideas more controllable and traceable to structured literature evidence.




\section{Experiments}

\subsection{Data}

We evaluate our method using a dataset derived from \textit{MAGenIdeas}~\cite{chen_enhancing_nodate}.
The dataset was constructed from ACL 2024 long papers, with metadata collected from ACL Anthology, OpenAlex, and Semantic Scholar.
The original collection contains 675 target papers, which are filtered by citation count, reference count, and author-information availability, resulting in 144 target papers.
For each target paper, the dataset provides the title and abstract, which are used as the basic input information in our experiments.

Following the evaluation practice of \textit{MAGenIdeas}, we construct a controlled evaluation split by randomly sampling 40 target papers from the 144-paper dataset.
For each target paper, all compared methods are provided with the same basic input, including the target paper title and abstract.
For retrieval-augmented baselines, we follow their original retrieval or context construction settings when available.
For \textsc{Graph2Idea}, the target paper is converted into a compact research profile for query generation, literature retrieval, knowledge graph construction, and graph-derived context-based idea generation.

\subsection{Experimental Setup}

\paragraph{Models and implementation.}
We use DeepSeek-v4-flash~\cite{deepseek2026v4} as the base LLM for target paper understanding, query generation, triple extraction, direction planning, idea synthesis, and automatic evaluation.
We use \textit{all-MiniLM-L6-v2}~\cite{wang2020minilmdeepselfattentiondistillation} as the embedding model for semantic similarity computation in graph-derived context construction.
For \textsc{Graph2Idea}, an LLM ranker selects 12 base papers from the retrieved candidate pool, and each base paper is expanded with its top 3 references according to keyword relevance, citation intent, publication year, abstract availability, and deduplication.
Each target paper is associated with up to four graph-derived contexts, each containing at most 40 serialized semantic triples.
Each context is paired with four prompt strategies; for each strategy, the LLM generates two directions, and each direction is used to synthesize one idea.
Thus, \textsc{Graph2Idea} generates up to 32 candidate ideas per target paper.
For fair comparison, all methods are configured to generate a comparable number of candidate ideas whenever possible.

\paragraph{Baselines.}
We compare \textsc{Graph2Idea} with four representative baselines: AI-Researcher~\cite{si_can_2024}, AI-Scientist~\cite{lu_ai_2024}, Future-Idea-Generation~\cite{kumar2025largelanguagemodelsunlock}, and MAGenIdeas~\cite{chen_enhancing_nodate}.
For AI-Researcher, AI-Scientist, and Future-Idea-Generation, we use their official open-source implementations without modifying the core generation pipelines.
For AI-Scientist, we evaluate only its research idea generation module and exclude downstream components such as experiment execution and paper writing.
For MAGenIdeas, we adopt its default setting with three discussion rounds and four agents, and initialize it with eight seed ideas to ensure a comparable number of generated ideas.

\paragraph{Metrics.}
Following prior automatic evaluation protocols, we assess generated ideas from three perspectives: \textit{Novelty}, \textit{Feasibility}, and \textit{Quality}.
Novelty follows the retrieval-grounded checking protocol used in AI-Scientist~\cite{lu_ai_2024}; feasibility follows the pairwise comparison setting of IdeaBench~\cite{guo2024ideabenchbenchmarkinglargelanguage}; and quality follows the Swiss-system tournament ranking protocol used in AI-Researcher~\cite{si_can_2024}.
All metrics are automatic proxy measures for early-stage idea assessment rather than definitive judgments of scientific validity.

For \textit{Novelty}, an evaluator generates a search query for each idea, retrieves the top-25 related papers, filters the 10 most relevant papers, and judges whether the idea is already covered by existing literature.
We report novelty as the proportion of ideas judged as novel.
For \textit{Relative Feasibility} and \textit{Relative Quality}, we use Swiss-system pairwise tournaments with LLM-based rankers.
The two metrics share the same comparison format but use different judging criteria.
For feasibility, the ranker focuses on implementability, data availability, experimental plausibility, and technical assumptions.
For quality, the ranker considers overall research value, including originality, significance, soundness, clarity, and potential impact.
Following prior settings, we report the proportion of ideas that win at least five comparisons for each metric.

\begin{table}[!t]
\centering
\scriptsize
\setlength{\tabcolsep}{5pt}
\renewcommand{\arraystretch}{1.08}
\caption{Overall performance comparison under the automatic evaluation protocol.}
\begin{tabular}{p{0.42\columnwidth}ccc}
\toprule
Method & Novelty $\uparrow$ & Rel. Quality $\uparrow$ & Rel. Feas. $\uparrow$ \\
\midrule
Future-Idea-Generation (2024) & 0.01 & 0.07 & 0.19 \\
AI-Scientist (2024) & 0.09 & 0.12 & 0.19 \\
MAGenIdeas (2025) & 0.20 & 0.19 & 0.06 \\
AI-Researcher (2025) & 0.45 & 0.24 & 0.22 \\
\midrule
\rowcolor[HTML]{CEEAF7}
\textbf{Graph2Idea (Ours)} & \textbf{0.52} & \textbf{0.29} & \textbf{0.28} \\
\bottomrule
\end{tabular}
\label{tab:main_results}
\end{table}
\subsection{Experimental Results}

Table~\ref{tab:main_results} reports the overall performance of different methods under the automatic evaluation protocol.
Compared with representative LLM-based idea generation baselines, our framework achieves the best performance across all three metrics, including novelty, quality, and feasibility.
In particular, our method obtains a \textit{Novelty} score of 0.52, outperforming the strongest baseline, AI-Researcher, which achieves 0.45.
This suggests that constructing structured graph contexts from retrieved literature can help the LLM generate ideas that are less likely to be judged as already covered by existing work.

Our method also achieves the highest \textit{Rel.quality} score of 0.29, compared with 0.24 for AI-Researcher and 0.19 for MAGenIdeas.
This indicates that decomposing generation into direction planning and idea synthesis can improve the overall competitiveness of generated ideas under pairwise LLM-based ranking.
More importantly, the \textit{Rel.feasibility} score of our method reaches 0.28, which is higher than all baselines.
This result suggests that the proposed framework does not improve novelty by producing overly speculative ideas; instead, the knowledge graph-based context provides structured evidence that helps maintain practical plausibility.

Overall, these results support the effectiveness of transforming retrieved literature from flat textual contexts into compact graph-based contexts.
By extracting relational knowledge from papers and using graph-grounded direction planning, our framework improves novelty while also maintaining strong feasibility and overall idea quality.

\section{Analysis and Discussion}
\subsection{Ablation Study}

We conduct ablation experiments to examine the contribution of direction generation, strategy-specific prompting, and graph-derived contexts.
The results are shown in Table~\ref{tab:ablation}.
In \textbf{w/o Direction}, we keep the graph-derived contexts and prompt strategies, but directly generate ideas without the intermediate direction planning stage.
In \textbf{w/o Strategy \& Direction}, we remove both multi-strategy prompting and direction generation, using a single general-purpose prompt with the graph-derived context.
In \textbf{w/o Graph Context}, we remove knowledge graph construction and graph-derived context construction, and directly provide the retrieved reference papers to the LLM as a flat textual context while keeping strategy- and direction-guided generation.

\begin{table}[!t]
\centering
\scriptsize
\setlength{\tabcolsep}{5pt}
\renewcommand{\arraystretch}{1.08}
\caption{Ablation study of the proposed framework.}
\resizebox{\columnwidth}{!}{%
\begin{tabular}{p{0.42\columnwidth}ccc}
\toprule
Method & Novelty $\uparrow$ & Rel. Quality $\uparrow$ & Rel. Feasibility $\uparrow$ \\
\midrule
w/o Direction & 0.47 & 0.28 & 0.27 \\
w/o Strategy \& Direction & 0.48 & 0.26 & 0.26 \\
w/o Graph Context & 0.50 & 0.28 & 0.27 \\
\midrule
\textbf{Graph2Idea (Full)} & \textbf{0.52} & \textbf{0.29} & \textbf{0.28} \\
\bottomrule
\end{tabular}
}
\label{tab:ablation}
\end{table}

The full method achieves the best performance across all three metrics.
Compared with \textbf{w/o Graph Context}, \textsc{Graph2Idea} shows consistent improvements in novelty, relative quality, and relative feasibility.
Although the improvements are modest, they suggest that graph-derived contexts provide a more useful structured representation of literature evidence than flat textual contexts.
Removing direction generation decreases all three metrics, indicating that explicit direction planning helps translate graph evidence into more concrete research ideas.
The \textbf{w/o Strategy \& Direction} variant further reduces quality and feasibility, suggesting that graph-derived contexts alone are not sufficient.
Instead, the results indicate that \textsc{Graph2Idea} benefits from the combination of structured graph contexts, strategy-specific prompting, and direction planning

\subsection{Case Study}

\begin{figure*}[t]
  \centering
  \includegraphics[width=\textwidth]{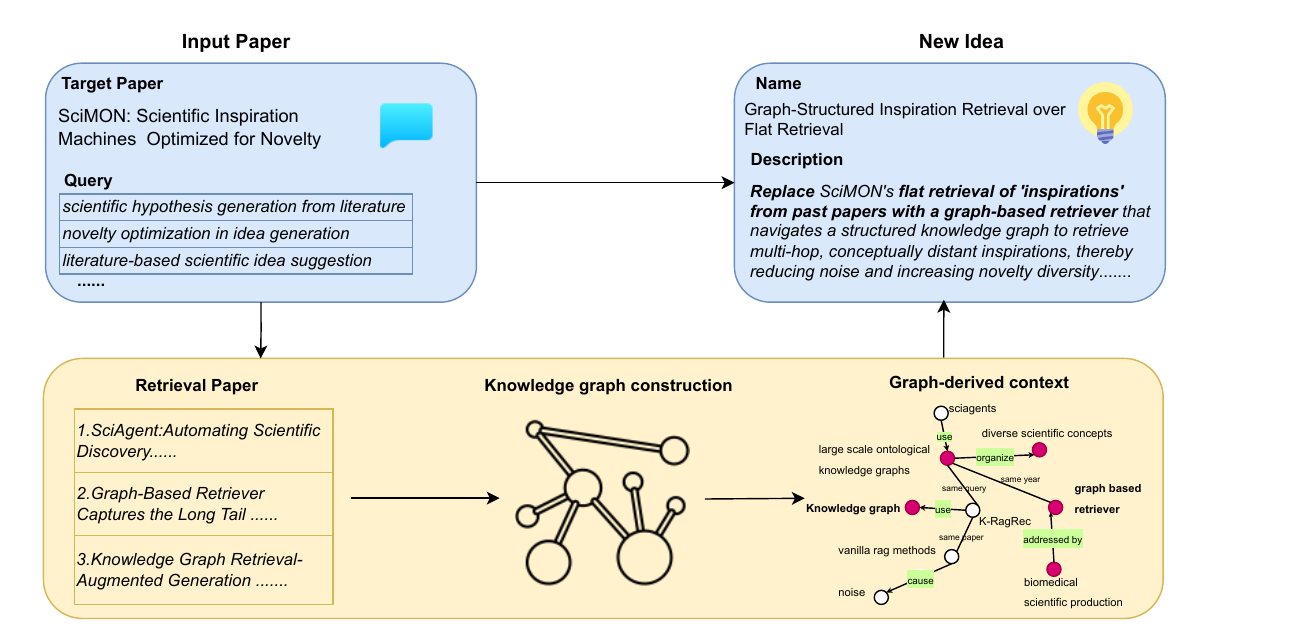}
  \caption{Case study on \textit{SciMON}. 
  \textsc{Graph2Idea} constructs graph-structured evidence from retrieved literature and generates a graph-based inspiration retrieval idea.}
  \label{fig:case_study}
\end{figure*}

To further illustrate how \textsc{Graph2Idea} uses graph-structured evidence for literature-grounded idea generation, we present a case study using \textit{``SciMON: Scientific Inspiration Machines Optimized for Novelty''} as the target paper.
As shown in Figure~\ref{fig:case_study}, the framework first identifies the target topic of scientific idea generation and novelty optimization.
It then retrieves related papers, including work on scientific agents, graph-based retrieval, and knowledge graph-enhanced retrieval-augmented generation.

After knowledge graph construction, the retrieved literature is organized into a graph-based evidence network rather than a flat list of papers.
This network illustrates the role of graph-derived relational contexts discussed earlier: it makes useful cross-paper relations more explicit while reducing dependence on long unstructured abstracts.
Semantic relations describe scientific knowledge, such as how knowledge graphs organize scientific concepts, how graph-based retrievers address long-tail knowledge, and how vanilla RAG may introduce noise.
Auxiliary structural links, such as publication year, source paper, or retrieval query, help connect otherwise sparse evidence nodes.
These auxiliary links are not treated as core scientific evidence in the serialized LLM context, but they help organize and visualize how retrieved papers and knowledge components are connected.

Based on this graph-based evidence, \textsc{Graph2Idea} generates the idea \textit{Graph-Structured Inspiration Retrieval over Flat Retrieval}.
The idea proposes to replace SciMON's flat retrieval of inspirations from prior papers with a graph-based retriever that navigates structured evidence to retrieve multi-hop and conceptually distant inspirations.
This generated idea is directionally similar to our framework because both emphasize graph-structured literature representations rather than flat textual contexts.
However, it is not a direct restatement of our method; instead, it applies the graph-structured perspective specifically to the inspiration retrieval stage of SciMON.
Moreover, the idea is judged as novel by the retrieval-grounded novelty evaluator.
This case suggests that \textsc{Graph2Idea} can recover meaningful graph-based methodological directions from retrieved literature and recombine them into new research ideas.

\section{Limitations}

Despite its effectiveness, the proposed framework has several limitations. First, the constructed knowledge graph depends on the quality of LLM-extracted triples. Scientific knowledge triples extracted from titles and abstracts are often sparse, heterogeneous, and weakly connected. To support graph organization and subgraph extraction, we use auxiliary information such as publication year, venue, field, and source-paper metadata. However, these auxiliary links are not always semantically informative for idea generation, and the final LLM context therefore focuses mainly on semantic knowledge triples. Developing more meaningful relation schemas and graph construction methods remains an important direction for future work. Second, our experiments are conducted on a subset of target papers derived from the \textit{MAGenIdeas} benchmark, mainly focusing on NLP papers from ACL 2024. Although this setting provides a controlled testbed, further validation across broader scientific domains is needed. In addition, the current assessment relies on automatic novelty checking, feasibility ranking, and quality ranking, without large-scale human expert evaluation. These automatic metrics should be interpreted as proxy evidence rather than definitive judgments of scientific validity, technical soundness, or long-term research potential. Incorporating expert review remains necessary for more reliable assessment of real-world scientific utility.




%
%
%
%
\bibliographystyle{splncs04}

\bibliography{reference}

\end{document}